\documentclass[final]{cvpr}

\usepackage{times}
\usepackage{epsfig}
\usepackage{graphicx}
\usepackage{amsmath}
\usepackage{amssymb}
\usepackage{diagbox}
\usepackage{soul}
\usepackage{subfig}
\usepackage[dvipsnames,table,xcdraw]{xcolor}
\usepackage{multirow}
\usepackage{hhline}
\usepackage{soul}

\newcommand{\gray}{\cellcolor[HTML]{C0C0C0}}

\definecolor{amber}{rgb}{1.0, 0.49, 0.0}

\usepackage[pagebackref=true,breaklinks=true,colorlinks,bookmarks=false]{hyperref}

\def\cvprPaperID{3383} % *** Enter the CVPR Paper ID here
\def\confYear{CVPR 2021}

\begin{document}

%%%%%%%%% TITLE
\title{Query2Label: A Simple Transformer Way to  Multi-Label Classification
}

\author{Shilong Liu$^{1,2}$, Lei Zhang$^{2}$, Xiao Yang$^{1}$, Hang Su$^{1}$, Jun Zhu$^{1}$\thanks{Corresponding author} \\
$^{1}$ Dept. of Comp. Sci. and Tech., BNRist Center, Institute for AI, Tsinghua-Bosch Joint ML Center\\
$^{1}$ Tsinghua University, Beijing, 100084, China  \hspace{2ex} 
$^{2}$ International Digital Economy Academy \\
\small{\{liusl20, yangxiao19\}@mails.tsinghua.edu.cn}, 
\small{leizhang@idea.edu.cn}, 
\small{\{suhangss, dcszj\}@mail.tsinghua.edu.cn} \\
}

\maketitle

%%%%%%%%% ABSTRACT
\begin{abstract}
This paper presents a simple and effective approach to solving the multi-label classification problem. The proposed approach leverages Transformer decoders to query the existence of a class label. The use of Transformer is rooted in the need of extracting local discriminative features adaptively for different labels, which is a strongly desired property due to the existence of multiple objects in one image. The built-in cross-attention module in the Transformer decoder offers an effective way to use label embeddings as queries to probe and pool class-related features from a feature map computed by a vision backbone for subsequent binary classifications. Compared with prior works, the new framework is simple, using standard Transformers and vision backbones, and effective, consistently outperforming all previous works on five multi-label classification data sets, including MS-COCO, PASCAL VOC, NUS-WIDE, and Visual Genome. 
Particularly, we establish $91.3\%$ mAP on MS-COCO.
We hope its compact structure, simple implementation, and superior performance serve as a strong baseline for multi-label classification tasks and future studies. 
The code will be available soon at https://github.com/SlongLiu/query2labels.

\end{abstract}

%%%%%%%%% BODY TEXT
%%%%%%%%% BODY TEXT
\section{Introduction}

% We study the multi-label classification problem in this paper. 
Multi-label image classification aims to gain a comprehensive understanding of objects and concepts in an image which has wide applications in realistic scenarios including image search, personal photo organization, digital asset management, medical image recognition~\cite{ge2018chest}, and scene understanding~\cite{shao2015deeply}.
\label{sec:2prob}
Compared with single label classification, multi-label classification requires special attention on two problems: 1) how to handle the label imbalance problem, and 2) how to extract features from region of interests. The former problem is because of the \emph{one-vs-all strategy}, i.e., 
it usually trains a batch of separate binary classifiers with each designed for recognizing a particular class,
which may lead to severely imbalanced numbers of positive and negative samples 
especially when the number of classes is large . 
The latter problem is because of the \emph{distributed objects}, i.e., an image often has multiple objects at different locations -- a globally pooled feature as normally used in single label classification may dilute the features and make it hard to identify small objects. 

It has witnessed significant attempts to solve the aforementioned issues. 
To balance positive and negative samples, many loss functions have been developed, such as focal loss~\cite{lin2017focal}, distribution-balanced loss~\cite{wu2020distribution}, and asymmetric loss~\cite{benbaruch2020asymmetric}. 
Some works have tried to address the second problem by utilizing spatial transformer network~\cite{wang2017multi}, adopting a global-to-local strategy~\cite{gao2020learning}, or using semantic label embeddings learned from label graph to discover the locations of discriminative features~\cite{you2020cross}. Compared with the first issue, solutions for the second problem are relatively less mature, requiring either specially designed network architectures or additional dependencies on label correlation.

\begin{figure}[t]
% \vspace{-1ex}
\begin{center}
  \includegraphics[width=1.0\linewidth]{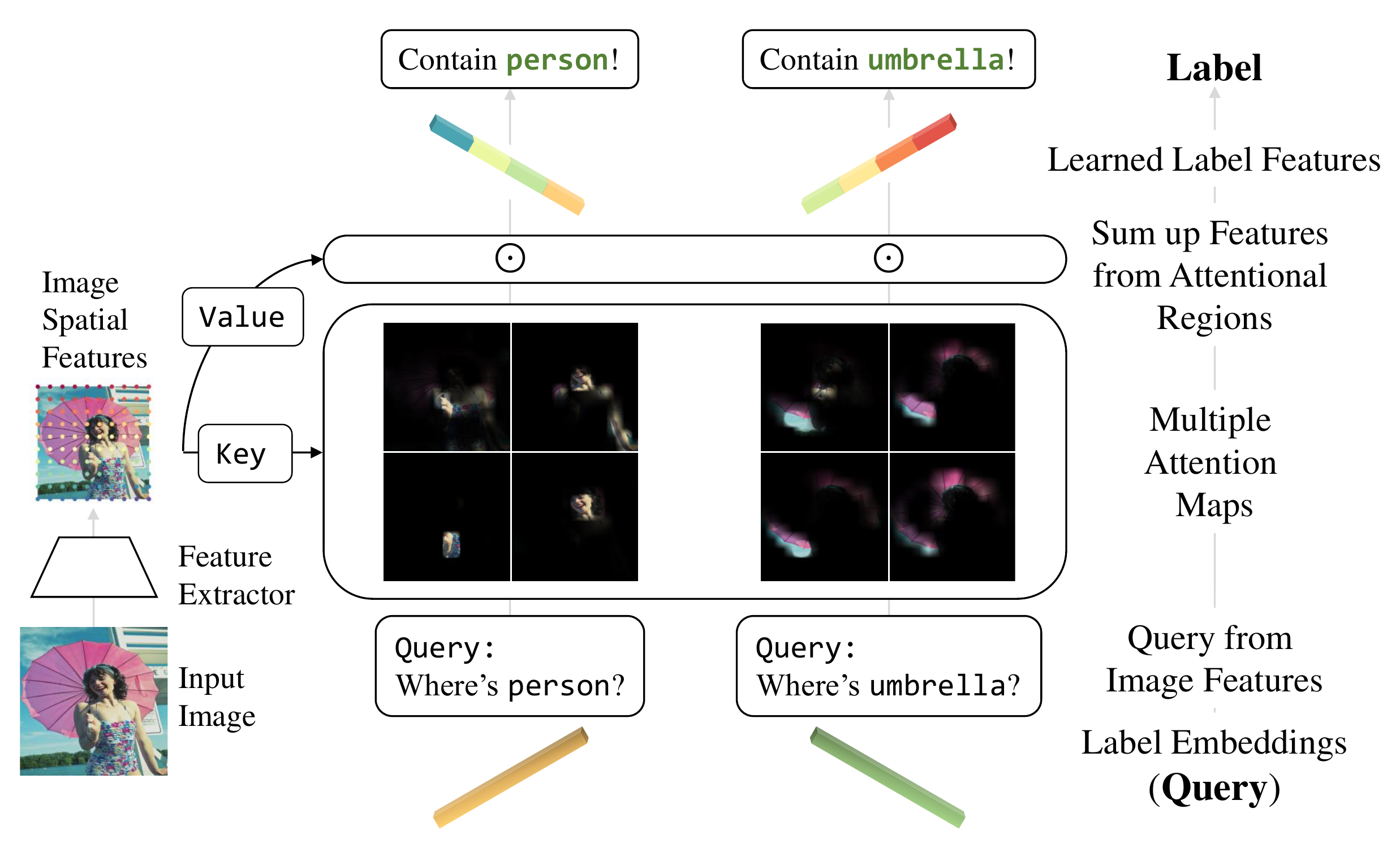}
\end{center}
% \vspace{-4ex}
  \caption{Illustration of Query2Label. Using cross attention for adaptively feature pooling through focusing on different parts (best view in colors).}
\label{fig:1fig}
% \vspace{-1ex}
\end{figure}

Motivated by the success of Transformer used in computer vision tasks~\cite{carion2020end, dosovitskiy2020image}, we present in this paper a simple yet effective solution using Transformer decoder to query the existence of a class label. We show that, without bells and whistles, the proposed solution leads to new state-of-the-art results and establishes strong baselines for its simple implementation and superior performance. We name the proposed solution as Query2Label and illustrate it in Fig. \ref{fig:1fig}. As shown in the figure, we use learnable label embeddings as queries to probe and pool class-related features via the cross-attention module in Transformer encoders. The pooled features are adaptive and more discriminative, leading to a superior multi-label classification performance.

The use of Transformer for solving multi-label classification is rooted in the need of extracting local discriminative features adaptively for different labels, which is a strongly desired property due to the existence of multiple objects in one image. While previous works \cite{zhu2017learning, benbaruch2020asymmetric} show that it is possible to use the globally average-pooled feature from the last layer of a convolutional neural network (CNN) for its simplicity in implementation, we argue that this will lead to inferior performance due to its discard of rich information in the convolutional feature map. The built-in cross-attention mechanism, which is called encoder-decoder attention in ~\cite{vaswani2017attention}, makes Transformer decoder a perfect choice for adaptively extracting desired features. By treating each label class as a query in a Transformer decoder, we can perform cross-attention to pool related features for the subsequent binary classification. The most related work to this idea is developed by You \etal ~\cite{you2020cross}, which, however, computes attention using cosine similarity with negative value clipping and uses the same feature for both key and value, greatly limiting its capability of learning locally discriminative features. 

Another advantage of Transformer is its multi-head attention mechanism, which can extract features from different parts or different views of an object class and thus is more capable of recognizing objects with occlusions and viewpoint changes. By contrast, the cross-modal attention in ~\cite{you2020cross} is merely a single-head attention, which is incapable of extracting features by parts or views. 

% conclusion of contributions
In this work, we develop a simple two-stage framework called Query2Label for multi-label classification by leveraging multi-layer Transformer decoders.
In the first stage, we use an image classification backbone 
to extract image features. The backbone could be either conventional CNN models such as ResNet~\cite{he2016deep} or recently developed Vision Transformer models~\cite{dosovitskiy2020image}. In the second stage, multiple Transformer decoder layers are leveraged, using label embeddings as queries to check the existence of each label by performing multi-head cross-attention to pool object features adaptively for subsequent binary  classification to predict the existence of the corresponding label.
Unlike ~\cite{you2020cross} in which the label embeddings are learned separately to take into account label correlations, we learn the label embeddings end-to-end to maximize the model potential and avoid the risk of introducing spurious correlations. The idea of using learnable label embeddings is inspired by DETR~\cite{carion2020end}. But the \texttt{queries} in DETR are class agnostic, whereas in our work each query (or label embedding) uniquely corresponds to one label class, making it more effective to extract class-related features. For this reason, in this paper, we will use \textit{label embedding} and \textit{query} interchangeably.

To handle the label imbalance problem, we adopt a simplified asymmetric loss~\cite{benbaruch2020asymmetric} by using different $\gamma$ values to weight positive and negative samples differently in focal loss. We found that this simple asymmetric loss works sufficiently well with this Transformer-based framework and leads to new state-of-the-art results on several multi-label benchmark data sets, including MS-COCO, PASCAL VOC, NUS-WIDE, and Visual Genome. 

Our contribution can be summarized as follows:

\begin{enumerate}
    \item We develop a simple Transformer-based two-stage framework Query2Label for multi-label classification, leading to an effective way to query the existence of a class label. 
    % that could extract object features adaptively with cross-attention and learn label features automatically from data, with transformer decoder added only. 
    To our best knowledge, this is the first time that the Transformer decoder architecture is used in classification. %The Transformer decoder architecture is standard, without any additional modifications. %While we do not claim any novelty in each block, 
    % We hope its simple implementation and superior performance will serve as a strong baseline for multi-label classification task and future studies.
    % , and the first time that a decoder is used in classification problem as well. 
    \item We show that, the built-in cross-attention module in Transformer decoders can adaptively extract object features and the multi-head attention further helps to decouple object representations into multiple parts or views, resulting in both improved classification performance and better interpretability. 
    % \item We show that, the multi-head attention could help to decouple objects into multiple parts and improve the performance of classification.
    % and it could be realized by multi-head attention layer of transformer decoder very well.
    % the multihead attention really matters here. With fixed feature length, a feature vector concentrated by vectors from multi-view is more expressive than that from single-view. Therefore, we suggest improving the number of heads while keep the feature length fixed.
    \item We verify the effectiveness of the proposed method with comprehensive experiments on several widely used data sets: MS-COCO, PASCAL VOC, NUSWIDE, and Visual Genome, and establish new state-of-the-art results on all these data sets.
\end{enumerate}

\section{Related Work}
\subsection{Multi-Label Classification} 
Multi-label classification task has attracted an increasing interest recently. The proposed methods can be categorized into three main directions as follows: 

\textbf{Improving loss functions.} As shown in Sec. \ref{sec:2prob}, one of the key concerns in multi-label classification is the imbalance of samples due to the use of one-vs-rest binary classifier for each category. Wu \etal~\cite{wu2020distribution} proposed a distribution-balanced loss to tackle the long-tailed problem by re-balancing weights and mitigating the over-suppression of negative labels. Ben-Baruch \etal~\cite{benbaruch2020asymmetric} proposed an asymmetric loss, which uses different $\gamma$ values to weight positive and negative samples in focal loss~\cite{lin2017focal}, and discarding easy negative samples by shifting the prediction probability. In our study, we adopt a simplified asymmetric loss which uses different $\gamma$ values for positive and negative samples without prediction probability shift. 

\textbf{Modeling label correlations.} For its nature of multi-labels on one image, the co-occurrence of concepts in a large-scale data set could be mined as prior knowledge for subsequent classification. 
Chen \etal~\cite{chen2019learning} proposed to learn category-correlated feature vectors by constructing a graph based on the statistical label co-occurrence and explored their interactions by performing neural graph propagation. 
Chen \etal~\cite{8784938} constructed a similar graph but based on class-aware maps, which is calculated by image level feature and classification weights, and constrained the graph by label co-occurrence.
% directly on label embeddings and generate classifiers that product with image features for classification. 
Rather than using static graph, Ye \etal~\cite{ye2020attention} updated static graph to dynamic graph by using a dynamic graph convolutional network(GCN) module for robust representations. While modeling label correlations can introduce additional gains in multi-label classification, it is also arguable that it may learn spurious correlations when the label statistics are insufficient. In our study, we directly learn label embeddings from data and encourage the network to focus on regions of interest to learn better feature representations and capture label relationships implicitly without graph networks.

\textbf{Locating regions of interest.} As an image normally has multiple objects, how to locate areas of interest becomes a concern in multi-label classification. Early methods~\cite{wei2015hcp, yang2016exploit} found proposals first and treated them as single-label classification. Wang \etal~\cite{wang2017multi} proposed to locate attentional regions corresponding to semantic labels by using a spatial transformer layer~\cite{jaderberg2015spatial} and predicted their scores with a Long Short-Term Memory  (LSTM) sub-network~\cite{hochreiter1997long}.
Gao \etal~\cite{gao2020learning} proposed a global-to-local discovery method to find proper regions with objects. All of these methods tried to find local regions to focus, but the discovered bounding boxes were coarse and often contained background information as well.
You \etal~\cite{you2020cross} computed cosine similarities between a label embedding and the feature map to derive an attention map after clipping negative values for class feature learning, which is a step forward. However, the cosine similarity with negative value clipping is likely to lead to a smoother and none spike attention, making it less effective in extracting desired local features (because it will cover larger areas than needed, see the visualized attention in Fig. 4 in \cite{you2020cross}). By contrast, we adopt the built-in cross-attention in Transformer as spatial feature selector to extract desired features, which is both simple and effective, thanks to the modularized design of Transformer and its readily available implementations in modern deep learning frameworks. 

\subsection{Transformer in Vision Tasks}
Transformer~\cite{vaswani2017attention} was first proposed to model long-range dependencies in sequence learning problems, and has been widely used in natural language processing tasks~\cite{devlin2018bert, lan2020albert, dai2019transformer, brown2020language, lagler2013gpt2, zhang2019ernie}. 
Recently, Transformer-based models have also been developed for many vision tasks
~\cite{dosovitskiy2020image, yuan2021tokens, srinivas2021bottleneck,carion2020end,zhu2020deformable,chen2020pre,guo2020pct, guo2021selfattention}
and shown great potentials. Chen \etal~\cite{chen2020generative} trained a sequence Transformer, named iGPT, to predict pixels auto-regressively. Dosovitskiy \etal~\cite{dosovitskiy2020image} proposed Vision Transformers (ViT), in which they split an image to multiple patches and feed them into a stacked Transformer architecture for classification. Carion \etal~\cite{carion2020end} designed an end-to-end object detection framework named DETR with transformer. Yuan \etal~\cite{yuan2021tokens} proposed Tokens-To-Token Vision Transformer (T2T-ViT) to address the patch tokenization problem. Srinivas \etal~\cite{srinivas2021bottleneck} replaced convolutional layers in last few ResNet Bottleneck~\cite{he2016deep} with Multi-Head Self-Attention and capture better global dependencies. More progress of applying Transformer in computer vision can be referred to \cite{han2020survey} and \cite{khan2021transformers}. 

Our approach also uses Transformers, but we leverage the built-in cross-attention in the Transformer decoder to locate object features for each label, which is largely different from most existing works ~\cite{dosovitskiy2020image, yuan2021tokens, srinivas2021bottleneck, cheng2021mltr} using the self-attention mechanism in Transformer encoders to improve feature representation. Our work is inspired by DETR~\cite{carion2020end}, but different in that the queries in DETR are class-agnostic and do not have clear semantics, whereas each query in our work uniquely corresponds to a semantic label.
% What's more, we leave the label features learning from data automatically, that makes the \texttt{query} semantic.

\section{Method}
Query2Label is a two-stage framework for multi-label classification, which uses Transformer decoders to extract features with multi-head attentions focusing on different parts or views of an object category and learn label embeddings from data automatically. In this section, we will present our framework first (Sec. \ref{sec:framework}), followed with a brief description to the loss function used (Sec. \ref{sec:loss}).

\begin{figure}[t]
% \vspace{-1ex}
\begin{center}
  \includegraphics[width=1.0\linewidth]{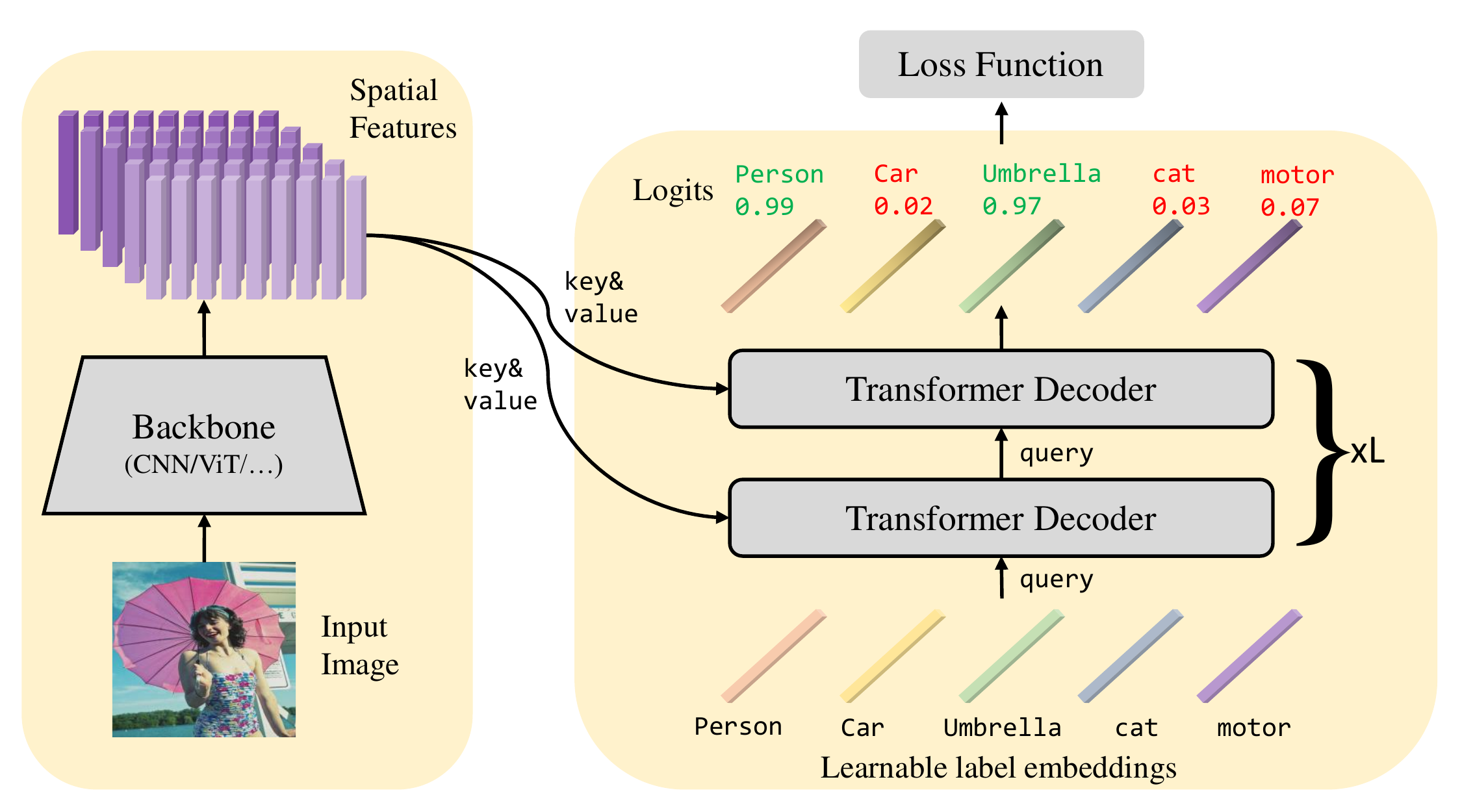}
\end{center}
% \vspace{-4ex}
  \caption{The framework of our proposed Query2Label (Q2L). After extracting spatial features of an input image, each label embedding is sent to Transformer decoders to \textit{query} (by comparing the label embedding with features at each spatial location to generate attention maps) and \textit{pool} the desired feature adaptively (by linearly combining the spatial features based on the attention maps). The pooled feature is then used to predict the existence of the queried label.}
\label{fig:framework}
\vspace{-1ex}
\end{figure}

\subsection{Framework}
\label{sec:framework}
Given an input image $x$, among a set of categories of interest, multi-label classification is to predict whether each category is present. A category could be either an object class (e.g. person, dog, table, etc.) or a scene category (grass, sky, etc). Assume there are $K$ categories in total and we denote the corresponding label of $x$ as $y=[y_1, ..., y_K]$, where $y_k\in\{0,1\}, k=1,...,K$, is a discrete binary indicator. $y_k=1$ if image $x$ has the $k$-th category label, otherwise $y_k=0$. Using $x$ as input, our model predicts the probabilities of the presence of each category, $p=[p_1,..., p_K]$, where $y_k\in[0,1], k=1,...,K$.

Fig. \ref{fig:framework} illustrates the framework of the proposed Query2Label (Q2L).
For an input image, it firstly feeds it into a backbone in the first stage to extract spatial features. The second stage is composed of two modules: a multi-layer Transformer decoder block for query updating and adaptive feature pooling, and a linear projection layer for computing prediction logits for each category. Note that our method is backbone-agnostic. That is, the second stage could be attached to any feature extractor. In this work, we mainly use convolutional neural networks as the feature extraction backbone, but Transformer-based networks such as ViT~\cite{dosovitskiy2020image} can also be used.

\vspace{0.2cm}
\noindent
\textbf{Feature extracting.}
Given an image $x\in \mathbb{R}^{H_0\times W_0\times 3}$ as input, we extract its spatial features $\mathcal{F}_0\in \mathbb{R}^{H\times W\times d_0}$ using the backbone, where $H_0\times W_0$, $H\times W$ represent the height and weight of the input image and the feature map respectively, and $d_0$ denotes the dimension of features.
%The backbone could be either convolutional neural networks~\cite{ciregan2012multi} or Transformer-based networks~\cite{dosovitskiy2020image}. 
After that, we add a linear projection layer to project the features from dimension $d_0$ to $d$ to match with the desired query dimension in the second stage and reshape the projected features to be $\mathcal{F}\in \mathbb{R}^{HW\times d}$. 

\vspace{0.2cm}
\noindent
\textbf{Query updating.}
After obtaining the spatial features of the input image in the first stage, we use label embeddings as queries $\mathcal{Q}_0\in \mathbb{R}^{K\times d}$ and perform cross-attention to pool category-related features from the spatial features using multi-layer Transformer decoders, where $K$ is the number of categories. We use the standard Transformer architecture, which has a self-attention module, a cross-attention module, and a position-wise feed-forward network(FFN). Each Transformer decoder layer $i$ updates the queries $\mathcal{Q}_{i-1}$ from the output of its previous layer as follows:

% \begin{equation}
\begin{alignat}{2}
    &\text{self-attn:} &&\mathcal{Q}_i^{(1)} = \text{MultiHead}(\widetilde{\mathcal{Q}}_{i-1}, \widetilde{\mathcal{Q}}_{i-1}, \mathcal{Q}_{i-1}), \cr
    &\text{cross-attn:}\quad &&\mathcal{Q}_i^{(2)} = \text{MultiHead}(\widetilde{\mathcal{Q}}_i^{(1)}, \widetilde{\mathcal{F}}, {\mathcal{F}}), \cr
    &\text{FFN:} &&\mathcal{Q}_i = \text{FFN}(\mathcal{Q}_i^{(2)}), 
\end{alignat}
\label{eq:decoder}
% \end{equation}

\noindent
where the tilde means the original vectors modified by adding position encodings,
$\mathcal{Q}_i^{(1)}$ and $\mathcal{Q}_i^{(2)}$ are two intermediate variables. Both the $\text{MultiHead(\texttt{query}, \texttt{key}, \texttt{value})}$ and $\text{FFN(}x\text{)}$ functions are the same as defined in the standard Transformer decoder~\cite{vaswani2017attention} and we omit their parameters for simplicity. As we do not need to perform auto-regressive prediction, we do not use attention masks. Thus the $M$ categories can be decoded in parallel in each layer.

Both the self-attention and cross-attention modules are implemented using the same $\text{MultiHead}$ function. The only difference is where \texttt{key} and \texttt{value} are from. In the self-attention module, \texttt{query}, \texttt{key} and \texttt{value} are all from label embeddings, whereas in the cross-attention module, \texttt{key} and \texttt{value} turn into the spatial features. The process of cross-attention can be described more intuitively: each label embedding $\mathcal{Q}_{{i-1},k} \in \mathbb{R}^{d}, k=1,...,K$ checks the spatial features $\widetilde{\mathcal{F}}$ where to attend and selects features of interest to combine. After that, each label embedding obtains a better category-related feature and updates itself. As a result, the label embeddings $\mathcal{Q}_0$ will be updated layer by layer and progressively injected with contextualized information from the input image via cross-attention.

Inspired by DETR, we treat the label embeddings $\mathcal{Q}_0$ as learnable parameters. In this way, the embeddings can be learned end to end from data and model label correlations implicitly. The difference between our approach and DETR is that our queries are class-specific and have clear semantic meanings, whereas the queries in DETR are class-agnostic and it is hard to predict which query will detect which category of objects.

\vspace{0.2cm}
\noindent
\textbf{Feature Projection.} Assuming that we have $L$ layers in total, we will get the queried feature vectors $\mathcal{Q}_{L}\in \mathbb{R}^{K\times d}$ for $K$ categories at the last layer. To perform multi-label classification, we treat each label prediction as a binary classification task and project the feature of each class $\mathcal{Q}_{L,k}\in \mathbb{R}^{d}$ to a logit value using a linear projection layer followed with a sigmoid function:
\begin{equation}
    p_k = \text{Sigmoid}\left(W_k^T \mathcal{Q}_{L,k}+b_k \right),
\label{eq:proj}
\end{equation}

\noindent
where 
$W_k\in\mathbb{R}^{d}$, 
$W=[W_1,...,W_K]^T\in\mathbb{R}^{K\times d}$, and $b_k\in\mathbb{R}$,
$b=[b_1, ...,b_K]^T\in\mathbb{R}^{K}$ are parameters in the linear layer, and $p=[p_1, ..., p_K]^T\in \mathbb{R}^{K}$ are the predicted probabilities for each category. Note that $p$ is a function which maps an input image $x$ to category prediction probabilities. $x$ is omitted for notation simplicity.

\subsection{Loss Function}
\label{sec:loss}
Thanks to the built-in cross-attention mechanism in Transformer decoders, our framework does not require a new loss function. Both the binary cross entropy loss and focal loss~\cite{lin2017focal} work well with our framework. To more effectively address the sample imbalance problem, we adopt a simplified asymmetric loss~\cite{benbaruch2020asymmetric}, which is a variant of focal loss with different $\gamma$ values for positive and negative values. In our experiments, we found it works the best. 

Given an input images $x$, we can predict its category probabilities $p=[p_1, ..., p_K]^T\in \mathbb{R}^{K}$ using our framework. Then we leverage the following asymmetric focal loss to calculate the loss for each training sample $x$:

\begin{equation}
\begin{aligned}
\mathcal{L}=\frac{1}{K}
\sum_{k=1}^{K}
    \begin{cases}
          (1-p_k)^{\gamma+}\log (p_k), & y_k=1,\\
          (p_k)^{\gamma-}\log (1-p_k), & y_k=0,\\
    \end{cases} 
\end{aligned}
\label{eq:loss}
\end{equation}

where 
$y_k$ is a binary label to indicate if image $x$ has label $k$. The total loss is computed by averaging this loss over all samples in the training data set $\mathcal{D}$. And the optimization is performed using stochastic gradient descent. 
By default, we set $\gamma+=0$ and $\gamma-=1$ in our experiments.

\section{Experiments}
\label{sec:exp}
To evaluate the proposed approach, we conduct experiments on several data sets, including MS-COCO~\cite{lin2014microsoft}, PASCAL VOC~\cite{everingham2015the}, NUS-WIDE~\cite{chua2009nus}, and Visual Genome~\cite{krishna2017visual}.
Following previous works, we adopt the average precision (AP) on each category and mean average precision (mAP) over all categories for evaluation. To better demonstrate the performance of models, we also present the overall precision (OP), recall (OR), F1-measure (OF1) and per-category precision (CP), recall (CR), F1-measure (CF1) for further comparison. See appendix for a more formal definition of these metrics.

\subsection{Implementation Details}
Unless otherwise stated, we will use the settings described below for all experiments. Following ASL~\cite{benbaruch2020asymmetric}, We adopt TResNetL~\cite{ridnik2020tresnet} as our backbone, as it performs better than the standard ResNet101~\cite{he2016deep} for this task under similar efficiency constraints on GPU. We resize all images to $H_0 \times W_0 = 448 \times 448$ as the input resolution and the size of the output feature from TResNetL is $H\times W\times d_0=14\times 14 \times 2432$. We set $d=d_0=2432$ in our experiments, hence the size of the final output feature in the first stage is $14\times 14 \times 2432$. The extracted features are fed into the second stage module after adding position encodings and reshaping. For the second stage, we utilize one Transformer encoder layer and two Transformer decoder layers for label feature updating. After the last Transformer decoder, we add a linear projection layer to compute logit predictions for all categories.

Note that the Transformer encoder is mainly used to further help fuse global context for better feature representation, but it can be removed for more efficient computation. In our experiments, our model works well even with only one Transformer decoder layer. See more ablations in Sec. \ref{sec:ablation}.

We leverage the ImageNet~\cite{deng2009imagenet} pre-trained model as our backbone, and update the whole model on the target multi-label classification data set. We trained the model for 80 epochs using the Adam~\cite{kingma2014adam} optimizer, with True-weight-decay~\cite{loshchilov2017decoupled} of $1e-2$, $(\beta_1, \beta_2)=(0.9, 0.9999)$, and a learning rate of $1\times 10^{-4}$. More implementation and training details are available in the supplementary materials.

\subsection{Comparison with State-of-the-art Methods}

\subsubsection{Performance on MS-COCO}
MS-COCO~\cite{lin2014microsoft} is a large-scale data set constructed for object detection and segmentation firstly and has been widely used for evaluating multi-label image classification recently. It contains $122,218$ images and covers $80$ common objects, with an average of $2.9$ labels per image. Notice that the mAP scores for MS-COCO are highly influenced by the input resolution. Thus we divide the results into two groups: medium resolution ($H,W\leq 600$) and high resolution($H,W>600$) and report them separately. 

For the medium resolution, we adopt our standard setting and report the comparison between our method and other state-of-the-art methods in Table \ref{Table:coco}. All the methods are evaluated in the resolution of $448\times 448$, except for SSGRL in $576\times 576$ and ADD-GCN in $512 \times 512$. Our proposed method outperforms all the previous methods in terms of mAP, OF1, and CF1, which are the most important metrics, as other metrics can be affected by the chosen threshold largely. 
In particular, our Q2L respectively outperforms ADD-GCN by $2.0\%$, SSGRL by $3.4\%$, and ASL by $0.6\%$. That demonstrates the superiority of our approach.

\begin{table*}[hbt!]
\setlength{\tabcolsep}{5pt}
\centering
\resizebox{\textwidth}{!}
{\begin{tabular}{c|c|c|c|c|c|c|c|c|c|c|c|c|c|c|c} 
\hline
\hline
\multicolumn{1}{c|}{\multirow{2}{*}{Method}} & 
\multicolumn{1}{c|}{\multirow{2}{*}{Backbone}} & 
\multicolumn{1}{c|}{\multirow{2}{*}{Resolution}} & 
\multicolumn{1}{c|}{\multirow{2}{*}{mAP}} & 
\multicolumn{6}{c|}{All}  & \multicolumn{6}{c}{Top 3} \\ 
\cline{5-16}
\multicolumn{1}{l|}{} &&&&CP & CR & CF1 & OP & OR & 
\multicolumn{1}{l|}{OF1} & CP & CR & CF1 & OP & OR & OF1 \\ 
\hline
\hline
SRN~\cite{zhu2017learning} & ResNet101 & 224$\times$224 & 77.1 & 81.6 & 65.4 & 71.2 & 82.7 & 69.9 & 75.8 & 85.2 & 58.8 & 67.4 & 87.4 & 62.5 & 72.9\\
ResNet-101~\cite{he2016deep}  & ResNet101 & 224$\times$224 &  78.3 & 80.2 &  66.7 &  72.8 & 83.9 & 70.8 & 76.8 & 84.1 & 59.4 & 69.7 & 89.1 & 62.8 & 73.6 \\
CADM~\cite{8784938} & ResNet101 & 448$\times$448  & 82.3 & 82.5 & 72.2 & 77.0 & 84.0 & 75.6 & 79.6 & 87.1 & 63.6 & 73.5 & 89.4 & 66.0 & 76.0 \\
ML-GCN~\cite{ML-GCN_CVPR_2019} & ResNet101 & 448$\times$448 & 83.0 & 85.1 & 72.0 & 78.0 & 85.8 & 75.4 & 80.3 & 87.2 & 64.6 & 74.2 & 89.1 & 66.7 & 76.3 \\
KSSNet~\cite{liu2018multi} & ResNet101 & 448$\times$448 & 83.7 & 84.6 & 73.2 & 77.2 & 87.8 & 76.2 & 81.5 & - & - & - & - & - & - \\
MS-CMA~\cite{you2020cross} & ResNet101 & 448$\times$448 & 83.8 & 82.9 & 74.4 & 78.4 & 84.4 & 77.9 & 81.0 & 86.7 & 64.9  & 74.3   & 90.9  & {67.2}  & 77.2  \\
MCAR~\cite{gao2020multi} & ResNet101 & 448$\times$448 & 83.8 & 85.0 & 72.1 & 78.0 & 88.0 & 73.9 & 80.3  & 88.1  & {65.5}  & {75.1}   & 91.0  & 66.3  & 76.7  \\ 
SSGRL~\cite{chen2019learning} & ResNet101 & 576$\times$576 & 83.8 & {89.9} & 68.5 & 76.8 & {91.3} & 70.8 & 79.7 & {91.9}  & 62.5 & 72.7 & 93.8 & 64.1 & 76.2  \\ 
C-Trans~\cite{lanchantin2020general} & ResNet101 & 576$\times$576  & 85.1 & 86.3 & 74.3 & 79.9 & \textbf{87.7} & 76.5 & 81.7 & 90.1 & 65.7 & 76.0 & 92.1 & \textbf{71.4} & 77.6  \\ 
ADD-GCN~\cite{ye2020attention} & ResNet101 & 576$\times$576 & 85.2 & 84.7 & 75.9 & 80.1 & 84.9 & \textbf{{79.4}} & 82.0 & 88.8  & {66.2}  & 75.8 & 90.3  & {68.5}  & 77.9  \\ 
\hline
Q2L-R101(Ours) & ResNet101 & 448$\times$448 &  {84.9} &  {84.8} &  {74.5}  &  {79.3} &  86.6 & 76.9 & 81.5 & 78.0 & 69.1 & 73.3 & 80.7 & 70.8 & 75.4  \\
Q2L-R101(Ours) & ResNet101 & 576$\times$576 & \gray \textbf{86.5} & \textbf{85.8} &  \textbf{76.7} & \gray \textbf{81.0} & 87.0 & 78.9 & \gray \textbf{82.8} & \textbf{90.4} & \textbf{66.3} &\gray \textbf{76.5} & \textbf{92.4} & 67.9 & \gray\textbf{78.3} \\
\hline
\hline
ASL~\cite{benbaruch2020asymmetric} & TResNetL & 448$\times$448 & {86.6} & {87.2} & {76.4} & {81.4} & {88.2} & {79.2} & {81.8}  & {91.8}  & 63.4  & {75.1} & {92.9} & 66.4 & {77.4}   \\
TResNetL~\cite{ridnik2021imagenet21k} & TResNetL(22k) & 448$\times$448 & {88.4} & - &  -  & -  & -  & - & - & - & - & - & - & - & - \\
\hline
Q2L-TResL(Ours) & TResNetL & 448$\times$448 &  {{87.3}} &  {\textbf{87.6}} & {76.5} &  {81.6} & \textbf{{88.4}} & {78.5} &  {83.1} & 
\textbf{{91.9}} & {66.2} &  {77.0} & \textbf{{93.5}} & {67.6} &  {78.5}  \\
Q2L-TResL(Ours) & TResNetL(22k) & 448$\times$448 &  \gray{\textbf{89.2}} &  86.3 & \textbf{81.4} & \gray\textbf{83.8} & 86.5 & \textbf{83.3} &\gray \textbf{84.9} & 91.6 & \textbf{69.4} & \gray\textbf{79.0} & 92.9 & \textbf{70.5} & \gray\textbf{80.2}  \\
\hline
\hline
MlTr-l~\cite{cheng2021mltr} & MLTr-l(22k) & 384$\times$384 & 88.5 &  86.0 &  81.4 &  83.3 &  86.5  & 83.4 & 84.9 & - & - & - & - & - & -\\
Swin-L~\cite{liu2021swin} & Swin-L(22k) & 384$\times$384 & 89.6 & \textbf{89.9} & 80.2 & 84.8 & \textbf{90.4} & 82.1 & 86.1 & 93.6 & 69.9 & 80.0 & 94.3 & 71.1 & 81.1 \\
CvT-w24~\cite{wu2021cvt} & CvT-w24(22k) & 384$\times$384 & 90.5 & {89.4} & 81.7 & 85.4 & 89.6 & 83.8 & 86.6 & 93.3 & 70.5 & 80.3 & 94.1 & 71.5 & 81.3\\
\hline
Q2L-SwinL(Ours) & Swin-L(22k) & 384$\times$384 &  {90.5} & {89.4} & 81.7 & 85.4 & {89.8} & 83.2 & 86.4 & \textbf{93.9} & 70.4 & 80.5 & \textbf{94.8} & 71.0 & 81.2\\
Q2L-CvT(Ours) & CvT-w24(22k) & 384$\times$384 & \gray {\textbf{91.3}} &  88.8 &  \textbf{83.2} &\gray \textbf{85.9} & 89.2 & \textbf{84.6} & \gray\textbf{86.8} & 92.8 & \textbf{71.6} &\gray \textbf{80.8} & 93.9 & \textbf{72.1} & \gray\textbf{81.6} \\
\hline
\hline
\end{tabular}}
\vspace{0.1cm}
\caption{Comparison of our method with known state-of-the-art models on MS-COCO at medium input resolution. The backbones noted with 22k are pretrained on the ImageNet-22k dataset.
Among them, mAP, OF1, and CF1 are the primary metrics (shaded in the table) as the others may be affected by the chosen threshold largely. All metrics are in \%.}
\label{Table:coco}
\end{table*}

For high resolution($640\times 640$) experiments, we adopt  TResNetXL~\cite{ridnik2020tresnet} as the backbone and remove the self-attention module in Transformer decoders for better training and inference efficiency. The results are shown in Table \ref{Table:coco_high_resolution}. Our method outperforms the best result in the literature and establishes a new state of the art. 

\begin{table}[hbt!]

\centering

\begin{tabular}{c|c|c|c} 
\hline

Method  & Architecture   & \begin{tabular}[c]{@{}c@{}}Input\\Resolution\end{tabular} & \begin{tabular}[c]{@{}c@{}}mAP\end{tabular}  \\ 
% \hline
% \hline
% \begin{tabular}[c]{@{}c@{}}
% ADD-GCN\\\cite{gao2020multi}
% \end{tabular}
%   & ResNet-101 GCN & 512 & 85.2 \\
\hline
% ASL     & TResNet-L & 448 & 86.5 \\
% ASL     & TResNet-L & 640 & 88.0 \\
ASL~\cite{benbaruch2020asymmetric}     & TResNetXL & 640$\times$640 & {88.4} \\
TResNet~\cite{ridnik2021imagenet21k}    & TResNetL(22k) & 640$\times$640 & {89.8} \\
\hline
Q2L-TResXL     & TResNetXL & 640$\times$640 & {89.0} \\
Q2L-TResL     & TResNetL(22k) & 640$\times$640 & \textbf{90.3} \\
\hline
\end{tabular}
% \medskip
\vspace{0.1cm}
\caption{Comparison of our method with ASL on MS-COCO for high input resolution of $640\times 640$. All metrics are in \%. }
\label{Table:coco_high_resolution}
\vspace{-0.2cm}
\end{table}

\subsubsection{Performance on PASCAL VOC}
PASCAL VOC 2007 and 2012~\cite{everingham2015the} are two frequently used data sets for multi-label classification. Each image in VOC contains one or multiple labels, corresponding to $20$ object categories.  In order to fairly compare with other methods, we follow the common setting to train our model on the \texttt{train-val} set and then evaluate its performance on the \texttt{test} set. Following previous works, we also pre-train the model on COCO for better performance.

\textbf{Results on VOC 07.} VOC 2007 contains $5,011$ images as the \texttt{train-val} set, and $4,952$ images as the \texttt{test} set. Results on VOC 07 are shown in Table \ref{table:voc07}. We can see that our method achieves the best mAP performance among all methods. 
We also observe the small margin between our results and ADD-GCN~\cite{ye2020attention}. In addition to the difference in input image resolution (ours $448\times 448$ and ADD-GCN's $512\times 512$), the small increase may indicate the limited data of VOC 07 and its saturated metric. 
Nevertheless, there might be some other factors that influence the results, as we outperform previous works on VOC 12 with a larger margin as shown in Table \ref{table:voc12}, whose results are reported by the evaluation server. We report results with ImageNet-1K pretrained backbone only in the main text for a fair comparison, and results with advanced backbones could be found in the appendix.

\textbf{Results on VOC 12.} VOC 2012 consists of $11,540$ images as the \texttt{train-val} set and $10,991$ images as the \texttt{test} set. Results on VOC 12 are shown in Table \ref{table:voc12}. As all the results are reported by its evaluation server, it is a much fairer comparison than a local test. Our method outperforms all other methods on all metrics with a large margin. 

\begin{table*}[t]
	\centering
	\resizebox{\textwidth}{!}{
		\begin{tabular}{|@{\,}c@{\,}| *{20}{@{\,}c@{\,}}|@{\,}c@{\,}|}
			\hline
			Methods   &aero &bike &bird &boat &bottle &bus &car &cat &chair &cow &table &dog &horse &mbike &person &plant &sheep &sofa &train &tv &mAP\\
			\hline
% 			\hline
			CNN-RNN~\cite{wang2016cnn}                &96.7 &83.1 &94.2 &92.8 &61.2 &82.1 &89.1 &94.2 &64.2 &83.6 &70.0 &92.4  &91.7 &84.2 &93.7 &59.8 &93.2 &75.3 &\textbf{99.7} &78.6 &84.0\\
			VGG+SVM~\cite{simonyan2015very}  &98.9 &95.0 &96.8 &95.4 &69.7 &90.4 &93.5 &96.0 &74.2 &86.6 &87.8 &96.0 &96.3 &93.1 &97.2 &70.0 &92.1 &80.3 &98.1 &87.0 &89.7\\
			Fev+Lv~\cite{yang2016exploit} &97.9 &97.0 &96.6 &94.6 &73.6 &93.9 &96.5 &95.5 &73.7 &90.3 &82.8 &95.4 &97.7 &95.9 &98.6 &77.6 &88.7 &78.0 &98.3 &89.0 &90.6\\
			HCP~\cite{wei2015hcp}  &98.6 &97.1 &98.0 &95.6 &75.3 &94.7 &95.8 &97.3 &73.1 &90.2 &80.0 &97.3 &96.1 &94.9 &96.3 &78.3 &94.7 &76.2 &97.9 &91.5 &90.9\\
            RDAL~\cite{wang2017multi}  &98.6 &97.4 &96.3 &96.2 &75.2 &92.4 &96.5 &97.1 &76.5 &92.0 &87.7 &96.8 &97.5 &93.8 &98.5 &81.6 &93.7 &82.8 &98.6 &89.3 &91.9\\
            RARL~\cite{chen2018recurrent} &98.6 &97.1 &97.1 &95.5 &75.6 &92.8 &96.8 &97.3 &78.3 &92.2 &87.6 &96.9 &96.5 &93.6 &98.5 &81.6 &93.1 &83.2 &98.5 &89.3 &92.0\\
            SSGRL~\cite{chen2019learning} (576)  &99.7 & 98.4  & 98.0  & 97.6  & 85.7 &  96.2 &  98.2 &  98.8 &  82.0 &  98.1  & 89.7 &  98.8  & 98.7  & 97.0  & 99.0  & 86.9  & 98.1  & 85.8  & 99.0 &  93.7 &  95.0\\
            % ASL(TResNetL)~\cite{benbaruch2020asymmetric} &-& - &-  &-  &- &- &- &- &-& -&-& - & - & - & - & - &- & - & - & - &{95.8} \\ 
            MCAR~\cite{gao2020learning} &	{99.7} &\textbf{99.0}& 98.5& {98.2} &  {85.4} & {96.9} &  {97.4} &  {98.9} &  {83.7} & 95.5 &  {88.8} &  {99.1} & 98.2& 95.1 &  {99.1} & 84.8 &  {97.1} &  {87.8} & 98.3 & {94.8} &{94.8} \\ 	
            ASL(TResNetL)~\cite{benbaruch2020asymmetric} & \textbf{99.9} &  98.4 & 98.9&  98.7&  86.8 & 98.2 & 98.7&  98.5&  83.1 & \textbf{98.3} & \textbf{89.5} & 98.8 & \textbf{99.2} & 98.6 & \textbf{99.3} & 89.5 & \textbf{99.4} & 86.8 & \textbf{99.6} & 95.2 & {95.8} \\ 
            ADD-GCN~\cite{ye2020attention} (576) & 99.8  &  \textbf{99.0}  &  98.4  &  \textbf{99.0}  &  86.7  &  98.1  &  98.5  &  98.3  &  \textbf{85.8}  &  \textbf{98.3}  &  88.9  &  98.8  &  99.0  &  97.4  &  99.2  &  88.3  &  98.7  &  \textbf{90.7}  &  {99.5}  &  \textbf{97.0}  &  96.0 \\
            \hline
            % \hline
            Q2L-TResL(Ours) & \bf{99.9} &  98.9 &  \textbf{99.0}  & {98.4}  & \textbf{87.7} &  \textbf{98.6} &  \textbf{98.8}  & \textbf{99.1} &  84.5 &  \textbf{98.3}  & {89.2}  & \textbf{99.2}  & \textbf{99.2} &  \textbf{99.2} &  \textbf{99.3}  & \textbf{90.2}  & {98.8}  & 88.3  & {99.5} &  95.5 &\textbf{96.1} \\ 
            
    \hline
	\end{tabular}}
% 	\vspace{0.05cm}
	\caption{Comparisons of our method with previous state-of-the-art methods on PASCAL VOC 2007, in terms of AP and mAP in $\%$. All results are reported at resolution $448\times 448$ except for the ADD-GCN and SSGRL, whose resolutions are noted in parentheses. Results with advanced backbones could be found in the appendix.}
    \label{table:voc07}
\end{table*} 
\begin{table*}[t]
	\centering
		
		\footnotesize
		\resizebox{\textwidth}{!}{
			\begin{tabular}{|@{\,}c@{\,}| *{20}{@{\,}c@{\,}}|@{\,}c@{\,}|}
				\hline
				Methods &aero &bike &bird &boat &bottle &bus &car &cat &chair &cow &table &dog &horse &mbike &person &plant &sheep &sofa &train &tv &mAP\\
				\hline
				VGG+SVM~\cite{simonyan2015very}&99.0 &89.1 &96.0 &94.1 &74.1 &92.2 &85.3 &97.9 &79.9 &92.0 &83.7 &97.5 &96.5 &94.7 &97.1 &63.7 &93.6 &75.2 &97.4 &87.8 &89.3\\			        
    			 Fev+Lv~\cite{yang2016exploit} &98.4 &92.8 &93.4 &90.7 &74.9 &93.2 &90.2 &96.1 &78.2 &89.8 &80.6 &95.7 &96.1 &95.3 &97.5 &73.1 &91.2 &75.4 &97.0 &88.2 &89.4\\
				HCP~\cite{wei2015hcp} &99.1 &92.8 &97.4 &94.4 &79.9 &93.6 &89.8 &98.2 &78.2 &94.9 &79.8 &97.8 &97.0 &93.8 &96.4 &74.3 &94.7 &71.9 &96.7 &88.6 &90.5\\
				% SSGRL*~\cite{chen2019learning}  &99.5 &95.1 &97.4 &96.4 &85.8 &94.5 &93.7 &{98.9} &86.7 &96.3 &84.6 &{98.9} &{98.6} &96.2 &98.7 &82.2 &{98.2} &{84.2}&98.1 &93.5 &93.9\\
				MCAR~\cite{gao2020learning}     &{99.6} &{97.1}&{98.3}&{96.6}&{87.0}&{95.5}&{94.4}&98.8&{87.0}&{96.9}&{85.0}&98.7&98.3&{97.3}&{99.0}&{83.8}&96.8&83.7&{98.3}&93.5  &{94.3}\\
				SSGRL~\cite{chen2019learning}(576)   & 99.7 &  96.1  & 97.7 &  96.5 &  86.9  & 95.8 &  95.0 &  98.9  & 88.3  & 97.6 &  87.4 &  99.1  & 99.2 &  97.3  & 99.0 &  84.8  & 98.3  & 85.8 &  99.2 &  94.1  & 94.8 \\
				ADD-GCN~\cite{ye2020attention}(576)  & 99.8 &  97.1 &  98.6 &  96.8 &  89.4 &  97.1 &  96.5 &  99.3 &  89.0  & 97.7  & 87.5 &  99.2  & 99.1  & 97.7 &  99.1 &  86.3 &  \textbf{98.8} &  87.0 &  99.3 &  95.4  & 95.5 \\
				\hline
				Q2L-TResL(Ours)   &\textbf{99.9} &\textbf{98.2}&\textbf{99.3}&\textbf{98.1}&\textbf{90.4}&\textbf{97.7}&\textbf{97.4}&\textbf{99.4}&\textbf{92.7}&\textbf{98.7}&\textbf{89.9}&\textbf{99.4}&\textbf{99.5}&\textbf{99.0}&\textbf{99.4}&\textbf{88.4}&\textbf{98.8}&\textbf{89.3}&\textbf{99.6}&\textbf{96.8}  &\textbf{96.6}\\
				\hline
	\end{tabular}}
% 	\vspace{0.05cm}
	\caption{Comparisons of AP and mAP in $\%$ of our model and state-of-the-art methods on PASCAL VOC 2012.  All results are reported at resolution 448$\times$448 except for the ADD-GCN and SSGRL, whose resolution is noted in parentheses. Different from VOC 07, results in VOC 12 are reported by the evaluation server.}
	\label{table:voc12}
\end{table*}

\subsubsection{Performance on NUS-WIDE}
NUS-WIDE~\cite{chua2009nus} is a real-world web image data set. It contains $269,648$ Flickr images and has been manually annotated with $81$ visual concepts. We follow the steps in \cite{benbaruch2020asymmetric} for evaluation, and compare the proposed method with previous state-of-the-art models in Table \ref{Table:nus}. 
As the resolutions of NUS-WIDE images are not high enough, we found the improvement of our method is not as significant as on MS-COCO and PASCAL VOC. But we still achieve a new state of the art on this data set. 

\begin{table}[hbt!]
\setlength{\tabcolsep}{5pt}
\centering
% \resizebox{\textwidth/2}{!}{
\begin{tabular}{c|c|c|c|l} 
\hline
Method & Backbone & mAP & CF1 & OF1   \\ 
% \hline
\hline
% S-CLs \cite{liu2018multi} & VGG16  & 60.1 & 58.7 & 73.7  \\
MS-CMA \cite{you2020cross} & ResNet101 & 61.4 & 60.5 & 73.8  \\
SRN \cite{zhu2017learning} & ResNet101  & 62.0 & 58.5 & 73.4  \\ 
ICME \cite{ML-GCN_CVPR_2019} & ResNet101  & 62.8 & 60.7 & 74.1  \\ 
Q2L-R101(Ours) & ResNet101 & \textbf{65.0} & \textbf{63.1} & \textbf{75.0} \\
\hline
Baseline~\cite{ridnik2020tresnet} & TresNetL    & 63.1 & 61.7  & 74.6   \\
Focal loss~\cite{lin2017focal} & TresNetL     & 64.0 & 62.9  & 74.7   \\
ASL~\cite{benbaruch2020asymmetric} & TresNetL    & {65.2} & {63.6}  & \bf{75.0}   \\
Q2L-TResL(Ours)& TresNetL  & \textbf{66.3} & \textbf{64.0}  & \textbf{75.0} \\
\hline
MlTr-l~\cite{cheng2021mltr} & MlTr-l(22k) & 66.3 & 65.0 & 75.8 \\
Q2L-CvT(Ours) & CvT-w24(22k) & \textbf{70.1} & \textbf{67.6} & \textbf{76.3} \\
\hline
\end{tabular}
% }
\vspace{0.1cm}
\caption{{Comparison of our methods to known state-of-the-art models on NUS-WIDE.} The backbones noted with 22k are pretrained on the ImageNet-22k dataset. All metrics are in \%.}
\label{Table:nus}

\end{table}

\subsubsection{Performance on Visual Genome}
Visual Genome~\cite{krishna2017visual} is a data set that contains $108,249$ images and covers $80,138$ categories. As most categories contain very few samples, \cite{chen2019learning} select images with the most frequent $500$ categories, and split the data set into train and test sets. They call the new data set VG500. We follow their setting and compare our model with prior methods in Table \ref{Table:vg}. For a fair comparison, we set the resolution of input images to $512\times 512$, and evaluate our method using both the ResNet-101~\cite{he2016deep} and TResNetL~\cite{ridnik2020tresnet} backbones. Although the previous 
state-of-the-art model SSGRL~\cite{chen2019learning}
use larger image resolution ($576\times 576$) than ours ($512\times 512$), our method outperforms all previous works and achieves a new SOTA on VG500. 
As the number of categories in VG500 is much larger than other data sets, it becomes more challenging for a simple spatially average-pooled feature to recognize all of the objects. Hence the advantages of our method are more obvious. The results indicate the importance and effectiveness of spatially adaptive feature attention in multi-label classification, particularly when the number of categories is large.

\begin{table}[hbt!]
\centering

\begin{tabular}{c|c} 
\hline
Method & mAP   \\ 
% \hline    
\hline
ResNet-101~\cite{he2016deep} & 30.9 \\
ResNet-SRN~\cite{zhu2017learning} & 33.5 \\
SSGRL(ResNet101)~\cite{chen2019learning} & 36.6 \\
C-Tran(ResNet101)~\cite{lanchantin2020general} & 38.4 \\
\hline
Q2L-R101(Ours) & 39.5\\ 
Q2L-TResL-22k(Ours) & \textbf{42.5}\\ 
\hline
\end{tabular}
\vspace{0.1cm}
\caption{Comparison of our method with prior state-of-the-art methods on VG500. All metrics are in \%. 
All results are reported at the input resolution of $576\times 576$.}
\label{Table:vg}
% \vspace{-0.5cm}
\end{table}

\subsection{Ablation Study}
\label{sec:ablation}

\textbf{Results on objects of different sizes.}
To further test Q2L's performance on objects of different sizes, we split the MS-COCO~\cite{lin2014microsoft} \texttt{test} set into three subsets for small objects, medium objects, and large objects respectively. Following the common definition, objects occupying areas less than and equal to $32 \times 32$ pixels are considered as “small objects”, less than and equal to $96\times 96$ pixels are marked as “medium objects”, and the others are “large objects”. We compare our Q2L with the baseline TResNetL model. The results are listed in Table \ref{Table:abla_size}. Our model outperforms baselines on all three categories, especially on medium objects.
The larger improvement on medium objects demonstrates the superiority of the spatially adaptive feature pooling, which helps collect information that may be diluted by average pooling. For small objects, although our method has made a big step forward, it remains a challenging problem to be solved, requiring finer-grained details to be extracted from images. 

\begin{table}[hbt!]
\centering

\begin{tabular}{c|c|c|c} 
\hline
Method  & small & medium  & large   \\ 
\hline
Baseline(TResNetL)~\cite{ridnik2020tresnet} & {37.8} & {74.2} & 84.2 \\
Ours(TResNetL+Q2L) & 39.5 & 77.5 & 86.1 \\
\hline
\end{tabular}
\vspace{0.1cm}
\caption{Comparison of improvement on objects with different sizes.}
\label{Table:abla_size}

\end{table}

\subsection{Visualization of Attention Maps}
\begin{figure}[t]
\begin{center}
  \includegraphics[width=0.9\linewidth]{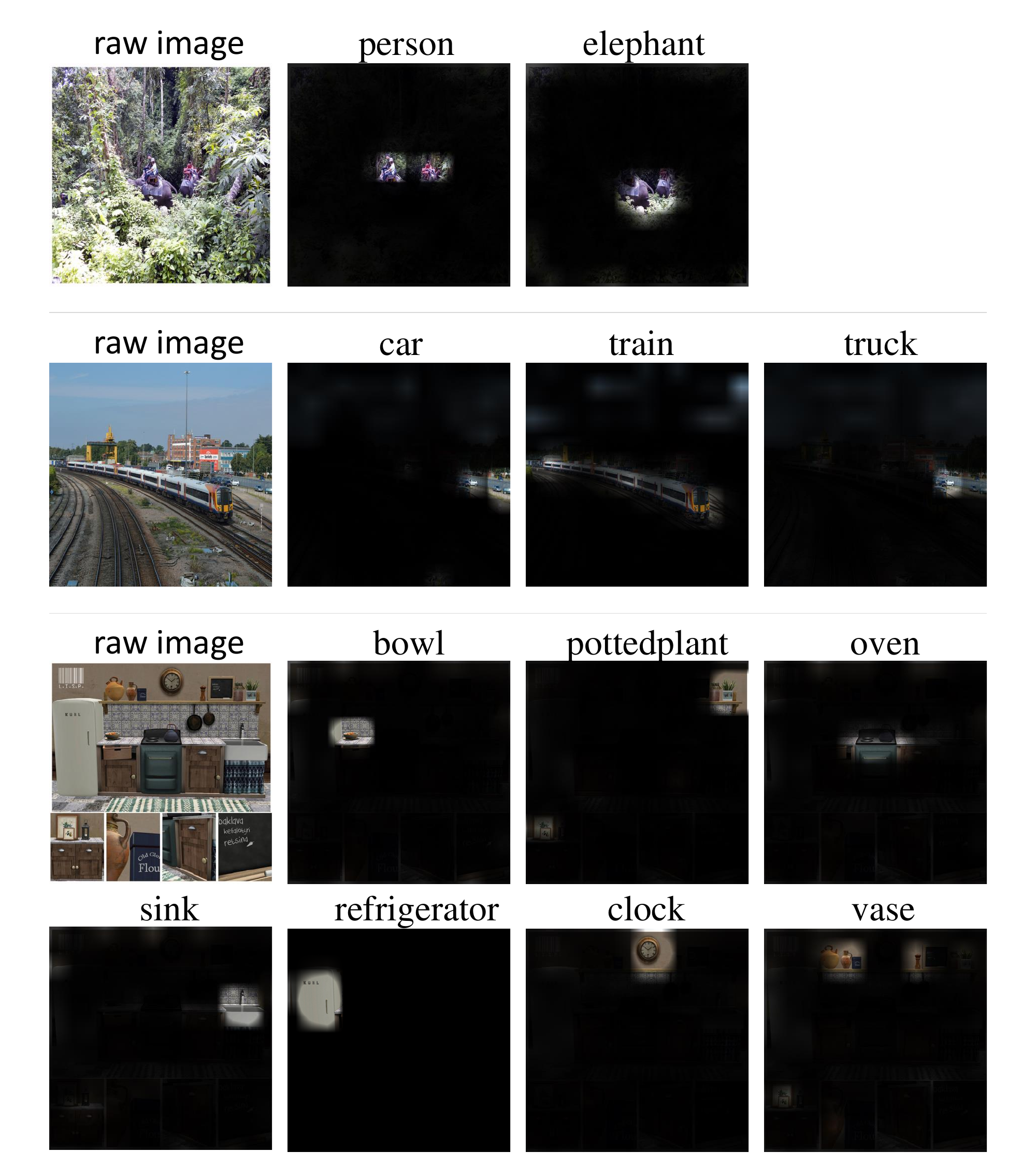}
\end{center}
  \caption{Visualization of cross-attention maps. 
  We plot the mean of each head's cross-attention maps, that represent similarities of a given \texttt{query} and extracted spatial features.
  Texts above images represent the ground truth labels (\texttt{query}) for the raw images. Best view in colors.
\vspace{-1ex}
}
\label{fig:attn_single}
\end{figure}

\begin{figure}[t]
% \vspace{-1ex}
\begin{center}
  \includegraphics[width=1.0\linewidth]{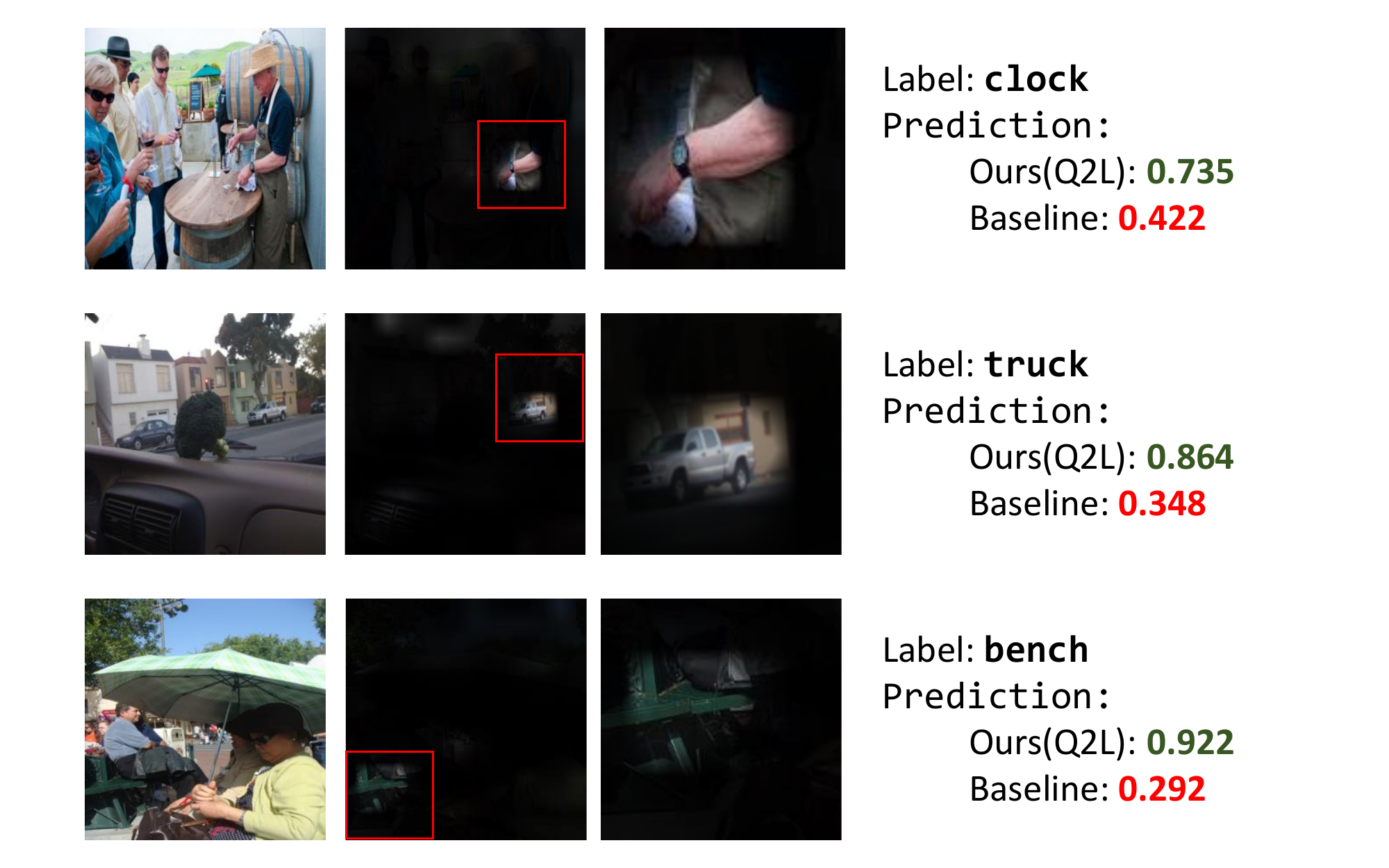}
\end{center}
% \vspace{-4ex}
\vspace{-0.2cm}
  \caption{Image examples classified correctly by Q2L but wrongly by the baseline TResNetL. The middle two columns are the mean attention maps of Q2L and the enlarged maps on focused regions respectively. The small scale of objects makes it difficult for TResNetL to recognize.  Best view in colors.
}
\label{fig:vis_compare}
\vspace{-0.2cm}
\end{figure}

To further demonstrate the effectiveness of cross-attention and adaptive pooling, we visualize some attention maps in Fig. \ref{fig:attn_single}. 
The attention map is analogous to the receptive field size in a raw image. We found that our model can locate the specified object approximately, especially on some small or medium objects. 
We also compare our Q2L model with baseline in Fig. \ref{fig:vis_compare}. 
It validates the effectiveness of the spatial adaptive pooling built into Transformer decoders and shows great potential for providing better interpretability.

\begin{figure}[t]
% \vspace{-1ex}
\begin{center}
  \includegraphics[width=1.0\linewidth]{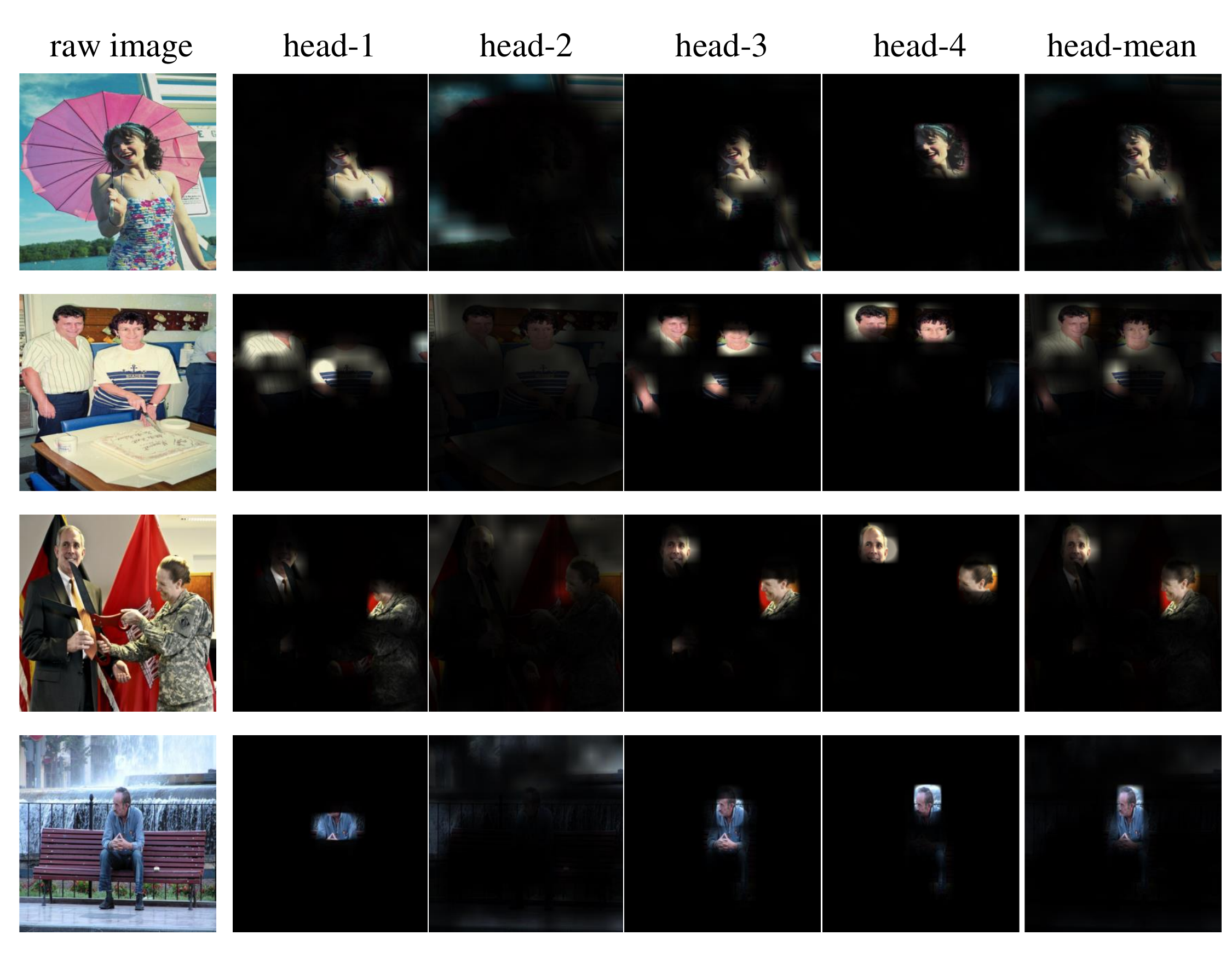}
\end{center}
% \vspace{-4ex}
\vspace{-0.2cm}
  \caption{Visualization of multi-head attention maps for the target label \texttt{person}. Each column in the middle represents an attention map for one head and the rightmost column averages the maps of all heads. Best view in colors.}
\label{fig:attn_multi}
\vspace{-0.1cm}
\end{figure}

Beyond single attention maps, we are also interested in finding out the role of multi-head attention in this task. For a given target \texttt{person}, we plot individual attention maps in each head and the mean attention map in Fig. \ref{fig:attn_multi}. We find that different heads are capable of focusing on different parts of targets.  For \texttt{person}, head-1, head-3, and head-4 focus on the shoulder, neck, and head respectively. The attention maps of head-2 are less informative, as there is no clear focus, which may indicate that head-2 is not utilized as the other three heads already collect sufficient information for classification. Focusing on different parts makes the model more robust to occlusion and view changes, and provides better interpretability for the superiority of our model. 

\section{Conclusion}
In this paper, we have presented a simple yet effective framework Query2Label (Q2L) for multi-label classification, which is developed based on Transformer decoders attached to an image classification backbone. The built-in cross-attention module in the Transformer decoder architecture offers an effective way to used label embeddings to query the existence of a class label and pool class-related features. The proposed framework consistently outperforms all prior works on several widely used data sets including MS-COCO, PASCAL VOC, NUSWIDE, and Visual Genome. We hope its simple model architecture and outstanding performance will serve as a strong baseline for future research on multi-label image classification.

\clearpage
\appendix

\section{More Implementation Details}
We adopt the official PyTorch implementation for both the backbone and Transformer modules~\cite{vaswani2017attention}. Following DETR~\cite{carion2020end}, we use 2D sine and cosine encodings to represent spatial positions. 
Each model was trained for $80$ epochs using Adam~\cite{kingma2014adam} and 1-cycle policy~\cite{smith2018disciplined}, with a maximal learning rate of $1e-4$. 
For regularization, we use Cutout~\cite{devries2017improved} with a factor of 0.5 and True-wight-decay~\cite{loshchilov2017decoupled} of $1e-2$. Moreover, we normalize input images with mean $[0,0,0]$ and std $[1,1,1]$, and use  RandAugment~\cite{cubuk2020randaugment} for augmentation.
Following common practices, we apply exponential moving average (EMA) to model parameters with a decay of $0.9997$. 
To speed up, we use mixed precision during model training. 
The entire code to reproduce the experiments will be made available.

\section{Metrics}
Beyond the average precision (AP) and mean average precision (mAP), we report more metrics in the experiments, including the overall precision (OP), recall (OR), F1-measure (OF1) and per-category precision (CP), recall (CR), F1-measure (CF1). These metrics are computed as follows:

\begin{equation}
\begin{aligned}
    &\mathrm{OP}=\frac{\sum_i M_c^i}{\sum_i M_p^i},  &&\mathrm{OR} = \frac{\sum_i M_c^i}{\sum_i M_g^i}, \\
    &\mathrm{CP}=\frac{1}{C}\sum_i\frac{M_c^i}{M_p^i},  &&\mathrm{CR} = \frac{1}{C}\sum_i\frac{M_c^i}{M_g^i}, \\
    &\mathrm{OF1}=\frac{2\times \mathrm{OP}\times \mathrm{OR}}{\mathrm{OP}+\mathrm{OR}},  &&\mathrm{CF1} = \frac{2\times \mathrm{CP}\times \mathrm{CR}}{\mathrm{CP}+\mathrm{CR}}, \\
\end{aligned}
\end{equation}

where $M_c^i$ is the number of images predicted correctly for the $i$-th category, $M_p^i$ is the number of images predicted for the $i$-th category,
and $M_g^i$ is the number of ground truth images for the $i$-th category.
For each image, we assign it a positive label if its prediction probability is greater than a threshold or negative otherwise. 
Note that these results may be affected by the chosen threshold. The $\mathrm{OF1}$ and $\mathrm{CF1}$ are the primary metrics among them, as they consider both recall and precision and are more comprehensive.

\section{Additional results on VOC07}
We show the results with ImageNet-1k pretrained backbones only in the maintext for a fair comparison on VOC 07. Additional results are listed in Table \ref{Table:voc07_add}.

\begin{table}[hbt!]
\centering

\begin{tabular}{c|c|c} 
\hline
Method & Resolution & mAP   \\ 
% \hline    
\hline
TResNetL & 448$\times$448 & 96.7 \\
Q2L-TResL(Ours) & 448$\times$448& 96.9 \\
Q2L-CvT(Ours) & 384$\times$384 & \textbf{97.3}\\ 
\hline
\end{tabular}
\vspace{0.1cm}
\caption{Comparison of our method with prior state-of-the-art methods on VOC07. Backbones in all models are pretrained on ImageNet-22k dataset. All metrics are in \%.}
\label{Table:voc07_add}
% \vspace{-0.5cm}
\end{table}

\section{More Visualization Results}
\begin{figure*}[t]
% \vspace{-1ex}
\begin{center}
  \includegraphics[width=1.0\linewidth]{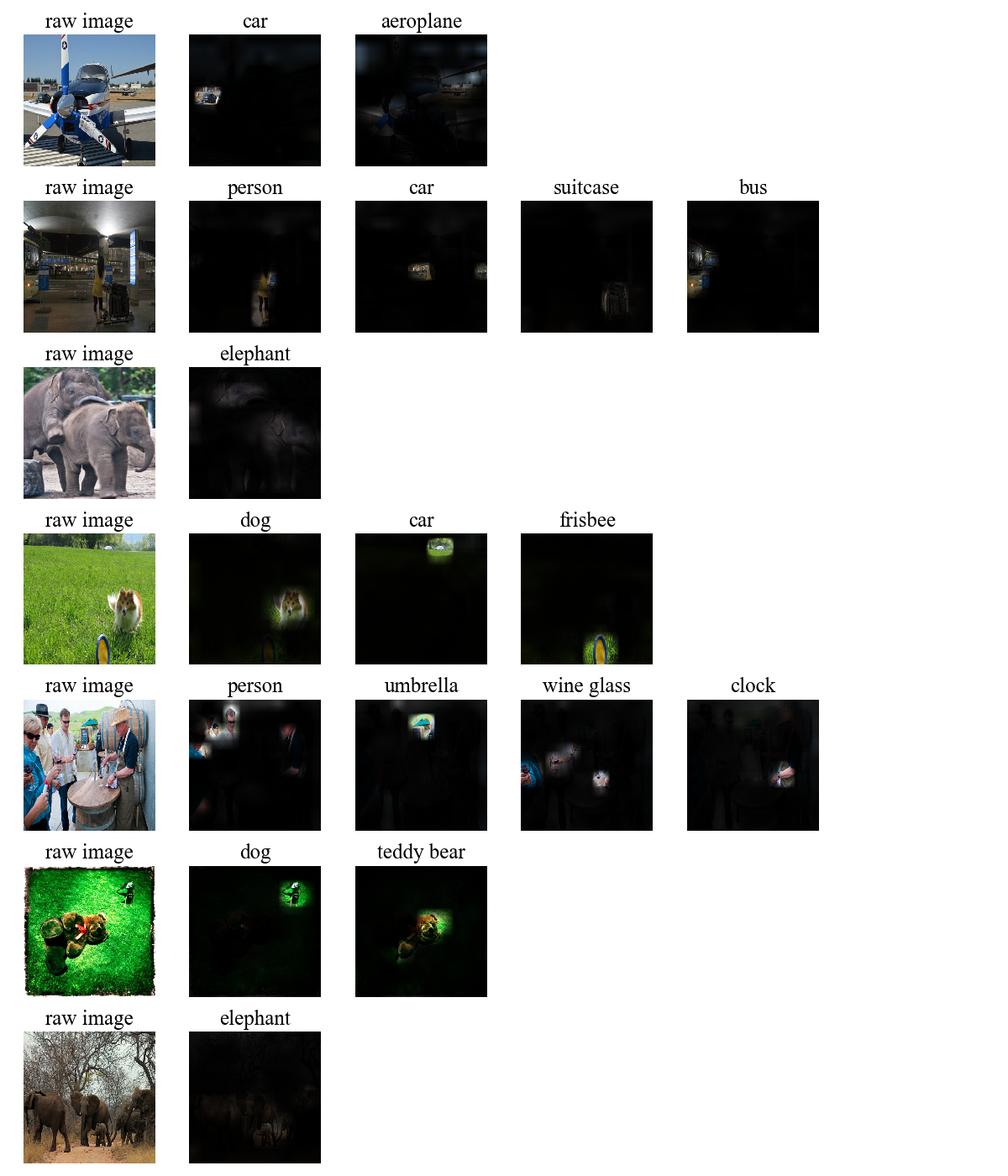}
\end{center}
    % \vspace{-4ex}
  \caption{More visualizations of cross-attention maps. 
  We plot the mean of each head's cross-attention maps, which represent similarities of a given \texttt{query} and the extracted spatial features.
  Texts above images represent the ground truth labels (\texttt{query}) for the raw images. Best view in colors.
% \vspace{-1ex}
}
\label{fig:single_attn_cont1}
\end{figure*}

\begin{figure*}[t]
% \vspace{-1ex}
\begin{center}
  \includegraphics[width=1.0\linewidth]{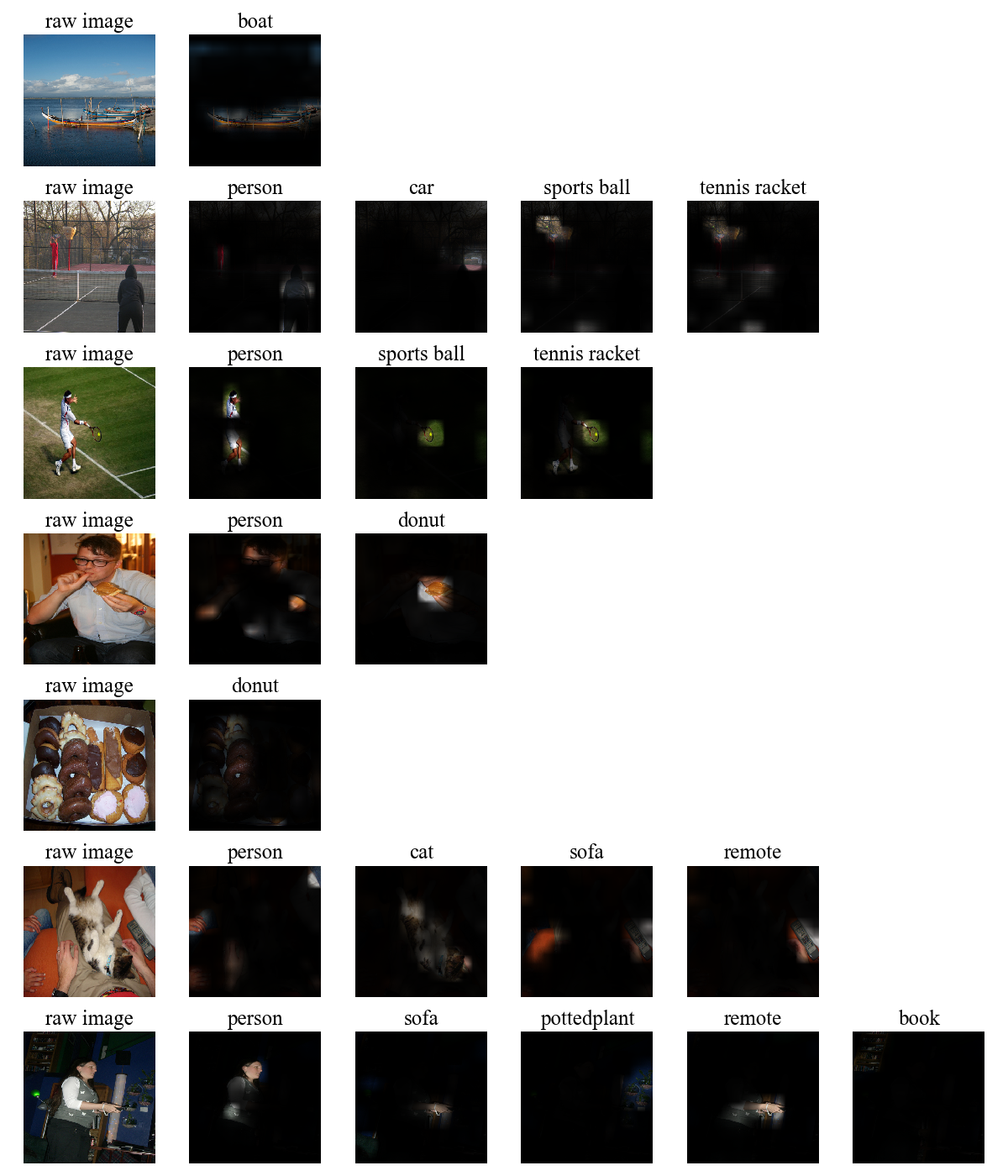}
\end{center}
    % \vspace{-4ex}
  \caption{More visualizations of cross-attention maps (continued). Best view in colors.
% \vspace{-1ex}
}
\label{fig:single_attn_cont2}
\end{figure*}

\begin{figure*}[t]
% \vspace{-1ex}
\begin{center}
  \includegraphics[width=1.0\linewidth]{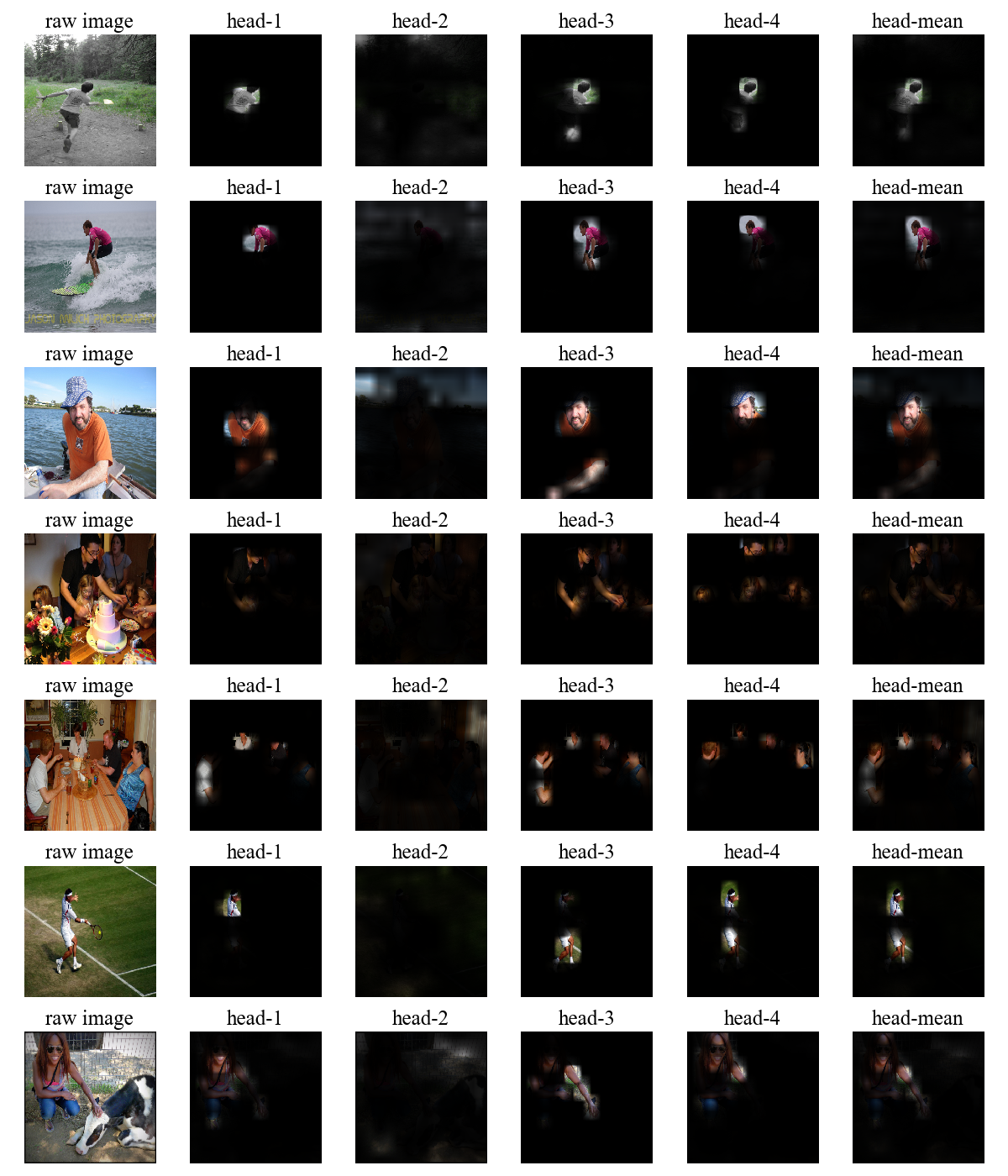}
\end{center}
    % \vspace{-4ex}
  \caption{More visualizations of multi-head attention maps for the target label \texttt{person}. Each column in the middle represents an attention map for one head and the rightmost column averages the maps of all heads. Best view in colors.
% \vspace{-1ex}
}
\label{fig:multi_attn_person}
\end{figure*}

We provide more visualization results of cross-attention maps on MS-COCO~\cite{lin2014microsoft}. 
To visualize the cross-attention maps, we compute the attention values between labels and pixels. Since each matrix adds up to $1$ and each value in the matrix is small, we divide the entire matrix by $0.06$ and clip it between $0$ and $1$ to get a better visualization result. Then for a target label, we resize its corresponding attention value matrices (for multiple attention heads) to the same size as raw images and use the resized matrices as the opacity of each pixel in images.
We plot the mean-head maps for more images in Fig. \ref{fig:single_attn_cont1} and Fig. \ref{fig:single_attn_cont2}. 
These figures provide an intuitive explanation for the effectiveness of our spatial adaptive pooling and the superiority of our method.
In addition, we show more multi-head attention maps of the target \texttt{person} in Fig. \ref{fig:multi_attn_person}.
% , and the target \texttt{dog} in Fig. \ref{fig:multi_attn_dog}. 
We find that different heads are capable of focusing on different parts of targets.

\clearpage

{\small
\bibliographystyle{ieee_fullname}
\bibliography{egbib}
}

\end{document}